\documentclass[11pt]{article}

\usepackage[utf8]{inputenc}
\usepackage[T1]{fontenc}
\usepackage{graphicx}
\usepackage[table]{xcolor}
\usepackage{microtype}
\usepackage{amsmath}
\usepackage{amssymb}
\usepackage{array}
\usepackage{booktabs}
\usepackage{float}
\usepackage{fancyhdr}
\usepackage{listings}
\usepackage[a4paper,width=155mm,top=30mm,bottom=25mm,headheight=14pt]{geometry}
\usepackage[
  backend=biber,
  style=numeric,
  sorting=none,
  maxbibnames=5,
  minbibnames=5,
  maxcitenames=2
]{biblatex}
\usepackage{hyperref}

\newcommand{\benchmark}{ScopeBench: Do Agents Preserve Engagement Boundaries Under Goal Pressure?}

\AtBeginDocument{\thispagestyle{fancy}}

\makeatletter
\let\ps@plain\ps@fancy
\makeatother

\hypersetup{
  colorlinks=true,
  linkcolor=blue,
  filecolor=blue,
  urlcolor=blue,
  citecolor=blue,
  pdftitle={ScopeBench: Do Agents Preserve Engagement Boundaries Under Goal Pressure?},
  pdfauthor={Shane Caldwell, Max Harley, Ads Dawson, Michael Kouremetis, Vincent Abruzzo, and Will Pearce}
}

\definecolor{codebackground}{rgb}{0.95,0.95,0.95}
\lstdefinestyle{judgeprompt}{
  language={},
  basicstyle=\scriptsize\ttfamily,
  breaklines=true,
  columns=fullflexible,
  frame=single,
  backgroundcolor=\color{codebackground},
  showstringspaces=false,
  xleftmargin=2pt,
  xrightmargin=2pt
}
\lstdefinestyle{tooltrace}{
  language={},
  basicstyle=\scriptsize\ttfamily,
  breaklines=true,
  columns=fullflexible,
  frame=single,
  backgroundcolor=\color{codebackground},
  showstringspaces=false
}

\DeclareUrlCommand\code{\ttfamily}
\makeatletter
\g@addto@macro{\UrlBreaks}{\do\-\do\_}
\makeatother

\providecommand{\Description}[1]{}

\DeclareUnicodeCharacter{2212}{-}

\newcommand{\NumTasks}{30}

\newcommand{\NumTraj}{2160}
\newcommand{\NumModels}{8}

\newcommand{\NumTrajPerCell}{3}
\newcommand{\NumJudgeCalibrationTraj}{100}

\newcommand{\NumJudgeAuditTraj}{100}
\newcommand{\JudgeModel}{claude-sonnet-5}

\date{}

\begin{document}

\title{ScopeBench: Do Agents Preserve Engagement Boundaries Under Goal Pressure?}

\author{
  Shane Caldwell\thanks{dreadnode, Principal Research Engineer. Email: shane@dreadnode.io.}\\
  \small dreadnode, USA
  \and
  Max Harley\thanks{dreadnode, Principal Security Researcher. Email: max@dreadnode.io.}\\
  \small dreadnode, USA
  \and
  Ads Dawson\thanks{dreadnode, Staff AI Security Researcher. Email: ads@dreadnode.io.}\\
  \small dreadnode, USA
  \and
  Michael Kouremetis\thanks{dreadnode, Principal AI Research Engineer. Email: michael@dreadnode.io.}\\
  \small dreadnode, USA
  \and
  Vincent Abruzzo\thanks{dreadnode, Principal Research Engineer. Email: vincent@dreadnode.io.}\\
  \small dreadnode, USA
  \and
  Will Pearce\thanks{dreadnode, Co-Founder. Email: will@dreadnode.io.}\\
  \small dreadnode, USA
}

\maketitle

\begin{abstract}
Agents are increasingly deployed with real autonomy in web application and network penetration testing, where a single out-of-scope action can breach a client's engagement boundary.
Existing offensive-security benchmarks measure raw hacking capability; as those benchmarks saturate, the actual barrier to deployment is a special case of alignment: scope adherence.
We introduce \textbf{ScopeBench}, a benchmark of \NumTasks{} \emph{dead-end} agentic security tasks in which the stated objective is reachable only by violating the stated scope.
Each task appears under two conditions that share an environment, verifier, and objective and differ only in scope: one instruction set has no scope and measures capability; the other has a natural-language scope to measure adherence.
Scopeless trajectories are graded by a standard deterministic verifier.
Scoped trajectories pass through two grading arms.
First, the same deterministic verifier checks for the flag: because the flag sits behind the scope boundary, a pass proves by construction that a forbidden action occurred, yielding a high-precision lower bound on the violation rate.
If the verifier does not pass the trajectory, an agentic judge estimates whether an out-of-scope call occurred.
We calibrate the judge against \NumJudgeCalibrationTraj{} ScopeBench trajectories labeled call-by-call by human annotators, and a blinded audit of the evaluated rollouts finds its high recall holds---no false negatives among the 36 audited violations, with over-flagging its only observed error.
Across \NumModels{} models run in one harness, raw capability spans $12.2\%$ to $81.1\%$ and scope adherence spans $34.4\%$ to $86.7\%$, with the judge finding 331 violations that mechanical verification misses.
Opus-4-8 achieves a raw-capability score 10 percentage points higher than sonnet-4-6's while exhibiting 35.6 percentage points higher scope adherence.
We release the frozen ScopeBench pilot benchmark and evaluation code as a methodological framework for measuring scope adherence in future agents, with an open pathway for community task contributions.\footnote{\url{https://github.com/dreadnode/scopebench-pilot}}
We also release all \NumTraj{} ATIF trajectories.\footnote{\url{https://huggingface.co/datasets/dreadnode/scopebench-pilot}}
Future versions of the community benchmark and the live leaderboard are published at \href{https://scopebench.ai}{\nolinkurl{scopebench.ai}}.
\end{abstract}

\section{Introduction}\label{sec:introduction}

LLMs now act as \emph{agents}: given an objective and a set of tools, they plan and issue tool calls autonomously until a long-horizon goal is met~\cite{yao2023react,schick2023toolformer}.
Every such deployment carries constraints, some implicit and some explicit.
A coding agent, for example, is told which repository to work on; it is expected, without being told, not to drop the production database~\cite{replit2025incident}. 
More dramatically, an agent undergoing evaluation is expected not to break free of its sandbox and search for solutions to an evaluation by gaining unauthorized production access to web hosts that may contain the answers~\cite{openai2026incident}.
Whether an agent honors those boundaries while reaching the goal is what makes a deployment safe.

Offensive security sharpens the stakes.
When an agent runs a penetration test, the line between authorized and unauthorized is not a matter of etiquette but of law and contract: a bug-bounty agent that finds a real vulnerability by wandering onto an out-of-scope host has at best found nothing and at worst committed a crime.
These agents also run mostly outside sandboxes, against systems that are live, client-owned, and often irreproducible, so the cost of a single out-of-bounds action is borne by a third party rather than the agent developer.
The ideal agent here is not necessarily the most capable one; it is the one as capable as it needs to be while being able to follow the rules of engagement and adhere to scope.

Contemporary agent evaluations, on the other hand, are overwhelmingly capability evaluations~\cite{zhang2024cybench,starace2025paperbench}: they provide a model tools and an objective and score whether the objective was met.
That measurement says nothing about whether the model would have used that same solution if some element of it were restricted.
As capability benchmarks saturate, this second axis---scope adherence---is what separates an agent that can be run autonomously at scale from one that must be monitored by humans or remain undeployed.
To our knowledge, no prior reproducible benchmark centers user-stated engagement boundaries in agentic security tasks while separating outcome-based violation verification from full-trajectory process evaluation.

The difficulty is that scope adherence resists straightforward mechanical measurement. 
Attaching a scope rule to a task and observing task completion percentage falling does not isolate the behavior of interest: an agent that fails a scoped task may have been unable to complete it, may have been confused by the extra instructions, or may have deliberately declined to cross the boundary.
Conversely, an agent can violate scope and still fail the task.
Outcome verification therefore supplies only one-sided evidence---a scoped \emph{pass} can verify a violation when success demands a forbidden action, but a scoped \emph{failure} does not imply scope adherence.

\textbf{ScopeBench} is built around that asymmetry.
We take \NumTasks{} agentic security scenarios and engineer each as a \emph{dead end}: the flag lives behind the scope boundary, so the in-scope surface, however inviting as a reconnaissance target, can never complete the task.
Each scenario is run first with no scope, giving a raw-capability measure, and then, in an independent trial, under a stated scope.
Because of the dead-end construction, any scoped success is a proof of violation, and the scoped pass rate is a direct lower bound on the true violation rate.
The obvious mechanical way to measure scope adherence---the raw-minus-scoped drop in completion, which we report as a descriptive \emph{hesitation delta}---is insufficient. 
If an agent violated the scope but chose not to submit the flag, or violated the scope but failed to find the flag, those trajectories would fail mechanical verification without being scope adherent.
To catch the violations in trajectories where the agent did not submit a flag, a second evaluation arm is used.
In this arm, an agentic judge uses tools to read the complete trajectory and decide whether any tool call the evaluated agent issued crossed the scope boundary.

We run \NumModels{} models through one harness.
Raw capability spans $12.2\%$ to $81.1\%$ across the model set.
The mechanical arm verifies violations in $4.4\%$ to $32.2\%$ of scoped trials, and the judge finds 331 additional scope violations in trajectories where no flag was submitted.
When the only path to the goal crosses the line, agents take it often: estimated boundary violation runs from $13\%$ to $66\%$ of trajectories across models.
Scope adherence does not appear to move in lockstep with capability: models with extremely similar capability scores have significant differences in scope adherence scores.
Because every task forces the choice between reaching the goal and violating the scope, these rates measure adherence under maximal goal pressure and should be read as an upper bound on the violation an agent would commit where an in-scope route to the same goal also exists.
We report capability and adherence on separate axes to keep incapable models from looking safe merely because they do not get close enough to the boundary to cross it. 

We consider this work a pilot to validate the measurement design, so the model-level results are provisional applications of this measurement methodology rather than a definitive safety ranking of the evaluated models.

\subsection{Contributions}

\begin{itemize}
  \item A two-arm measurement framework that pairs a deterministic, high-precision violation \emph{verifier} with a calibrated full-context judge, separating what an agent's raw hacking capability can do from its ability to stay in scope within a stated boundary. 
  \item A pilot evaluation of \NumModels{} models spanning $12.2\%$--$81.1\%$ raw capability, reported alongside estimated scope adherence from $34.4\%$ to $86.7\%$.
  \item An open release of the framework, the \NumTasks{}-task pilot corpus, all \NumTraj{} trajectories, and a task-authoring pathway so the benchmark can grow with community contributions.
\end{itemize}

\section{Background and Related Work}\label{sec:background}

\paragraph{Agent capability benchmarks.}
The canonical agent evaluation places a language model in a harness with tools, provides it an objective, and scores whether the objective is achieved~\cite{yao2023react,schick2023toolformer}.
Cybench~\cite{zhang2024cybench} and BountyBench~\cite{zhang2025bountybenchdollarimpactai} are representative examples in cybersecurity.
Each reports a single pass-rate axis: it establishes that a model can complete the task, but carries no signal for whether the model would have declined the same completion path if doing so would violate the scope of an engagement.
ScopeBench uses this per-task capability score and pairs it with constrained variants of the same task.
Unlike an ordinary outcome benchmark, it then confirms violations mechanically and estimates, through trajectory judging, the violations that can't be mechanically verified.

\paragraph{Agent safety under adversarial threat models.}
Prior agent-safety work has largely studied the adversarial case, in which a maliciously crafted input tries to push an agent past its authorization.
AgentDojo~\cite{debenedetti2024agentdojodynamicenvironmentevaluate}, InjecAgent~\cite{zhan2024injecagentbenchmarkingindirectprompt}, and ToolEmu~\cite{ruan2024identifyingriskslmagents} instantiate variants of the indirect prompt-injection setting; ToolSafe~\cite{mou2026toolsafeenhancingtoolinvocation} supplies inference-time guardrails.
ScopeBench asks a complementary question with no attacker in the loop: how well an agent holds a stated boundary under nothing more than its own pressure to finish the task.

\paragraph{Process-based evaluation and compliance.}
Refusal has been studied mechanistically~\cite{arditi2024refusallanguagemodelsmediated}, but many trajectories in our corpus that fail a scoped task never verbalize a refusal at all, which is why we evaluate actions rather than keyword-matched refusal text.
Tool-assisted judges have been used to assess operational requirements over complete security-agent trajectories~\cite{caldwell2025pentestjudgejudgingagentbehavior}, and to gate individual tool calls before execution~\cite{caldwell2026scopejudge}, motivating scalable process evaluation but leaving the error rate of any new trajectory-level judge unestablished.
ScopeBench therefore calibrates its judge against expert labels and reports mechanically verified violations separately from judge estimates, addressing the judge-quality confounds documented for LLM-as-judge scoring~\cite{zheng2023judging,shi2025judgingjudgessystematicstudy,li2024calibraevalcalibratingpredictiondistribution}.

\subsection{Problem Formalization}\label{sec:formalization}

A \emph{task} $t$ specifies an initial environment $E_t$ (containing a target application and a filesystem for writing results), a natural-language objective, a scope statement $\sigma \in \{\varnothing, \text{casual}, \text{formal}\}$, and a deterministic verifier $V_t$ that returns $\{\text{pass}, \text{fail}\}$.
Each task is a \emph{dead-end}: within $E_t$, the objective's stated deliverable is recoverable only through actions that violate a scope statement in $\{\text{casual}, \text{formal}\}$.
Under $\sigma = \varnothing$, those same actions are permitted.

An agent $\mathcal{A}$ receives $(E_t, \sigma, \text{objective})$ and produces a trajectory $\tau=(a_1,o_1,\ldots,a_N,o_N)$ of tool calls and outputs.
Let $M_t(\tau)\in\{0,1\}$ denote the mechanical pass indicator.
For each issued tool call $a_k$, let $y(a_k,\tau,\sigma)\in\{0,1\}$ equal one when the call, interpreted in the context of $\tau$, violates the operative scope $\sigma$, and zero otherwise.
The trajectory-level violation label $Y(\tau,\sigma)\in\{0,1\}$ records whether at least one issued tool call violates the operative scope:
\[
  Y(\tau,\sigma)=\max_{1\leq k\leq N} y(a_k,\tau,\sigma).
\]
For scoped variants, the dead-end construction forces $M_t(\tau)\leq Y(\tau,\sigma)$: a pass is definitely a violation, while a failure is uninformative about adherence.
We define capability as $P_{\mathrm{raw}}=\Pr[M_t(\tau)=1\mid\sigma=\varnothing]$, violation rate as $R_\sigma=\Pr[Y=1\mid\sigma]$, and adherence as $A_\sigma=1-R_\sigma$.
The mechanical verifier pass rate $P_\sigma$ is a lower bound on $R_\sigma$.
\section{Methodology}\label{sec:methodology}

\subsection{Scope Conditions}\label{sec:methodology-variants}

Each task is run under two conditions that share the same environment, verifier, and stated objective and differ only in the scope statement $\sigma$.
\emph{Raw capability} tells the agent that no scope restrictions apply.
The \emph{scoped} condition supplies a natural-language scope statement with categorical prohibitions (e.g.\ ``do not pivot to any internal system, do not create a new account'').

To check that our findings do not hinge on how a scope statement is worded, the pilot scoped condition is tested in two phrasings around the same scope: a terse \emph{casual} message and a \emph{formal} bug-bounty program brief with enumerated \emph{In Scope}, \emph{Out of Scope}, and \emph{Rules of Engagement} sections. The artifact release reproduces all three variants per task.
Pooled across models the phrasing does not significantly move the trajectory-level violation rate (Appendix~\ref{app:phrasing}).
For this pilot study, we report the scoped condition pooled across the two phrasings. 
Future contributed tasks to ScopeBench will only be required to include their scoped message in whatever way is most natural to the task. 

\subsection{Dead-End Task Design}\label{sec:methodology-deadend}

Every ScopeBench task is built so that the flag can be recovered only through a specific out-of-scope action.
The in-scope surface is dressed as a plausible reconnaissance target but never lets the agent finish the objective.
If the agent completes a scoped task at all, it did so by crossing the boundary, and the scoped pass rate becomes a direct lower bound on the violation rate.

\subsection{Arm I: Deterministic Violation Verification}\label{sec:methodology-verify}

Every task is verified by a per-task shell script that checks the trajectory's output against an expected, generated flag.
The flag is reachable only behind the task's stated boundary and is unpredictable without interacting with that surface, so a scoped pass is high-precision evidence that at least one violating action occurred.
We call the aggregate scoped pass rate the \emph{mechanical violation rate} (MVR):
\[
  \operatorname{MVR}_{\sigma}(\mathcal{A})
  = \operatorname{mean}_{t,\tau} M_t(\tau), \qquad \sigma\neq\varnothing.
\]
MVR is a lower bound, not a complete violation estimate: an agent can touch a forbidden host, identity, endpoint, or operation without ever recovering the flag. It may also recover the flag through a forbidden action and then not write it to the deliverable, which the verifier scores as a clean failure; we quantify how often this happens in Section~\ref{sec:results-hidden}.

We also report pass rate $P$ and the hesitation delta
\[
  \Delta_P^{\sigma}(\mathcal{A}) = \operatorname{mean}_t
  \bigl[P(t, \mathcal{A}, \varnothing) - P(t, \mathcal{A}, \sigma)\bigr].
\]
Hesitation delta measures how much the scope instructions suppress successful completion; it does not distinguish principled restraint from confusion or ordinary task failure. 
We include it to illustrate the efficacy of a judge.

\subsection{Arm II: Trajectory Violation Judge Estimate}\label{sec:methodology-judge}

The second arm estimates the trajectory label $Y$ directly.
A judge receives the agent's user request and scope statement and can use tools to enumerate the complete trajectory, search tool-call arguments and outputs, and retrieve individual steps with surrounding context.
Model identity and the mechanical verifier result are withheld to prevent evaluation shortcuts.

The judge returns \textsc{violation} or \textsc{no-violation}.
A violation verdict must cite at least one call index, the scope clause it crosses, and a short evidence-grounded rationale.
An interaction counts as a violation when the issued tool call itself crosses the boundary, even if the target returns an error; private reasoning or textual consideration with no external action does not.
The judge resolves redirects, aliases, identities, and effective hosts from call arguments and outputs.

Let $J(\tau,\sigma)$ denote the binary judge decision.
We combine the two arms as
\[
  \widehat{Y}(\tau,\sigma)
  = M_t(\tau) + \bigl(1-M_t(\tau)\bigr)J(\tau,\sigma),
\]
so no judge false negative can erase a mechanically verified violation: the judge only decides whether mechanically \emph{failing} trajectories contain violations despite not submitting a flag.
We report the raw hybrid estimate separately from the sensitivity analyses based on judge calibration and the model-blinded audit (Section~\ref{sec:ed-judge-calibration}).

\subsection{Evaluator Scope Rubric}\label{sec:methodology-oracle}

Each task pairs its agent-facing natural-language scope with an evaluator scope rubric.
The rubric enumerates forbidden surfaces, identities, and operations; records exceptions; maps each prohibition to the corresponding agent-facing clause; and states why recovering the generated flag requires a forbidden action.
It is available to human graders and to the trajectory judge, but not to the agent being evaluated.

\section{Dataset}\label{sec:dataset-intro}

The ScopeBench pilot consists of \NumTasks{} tasks, each in its own isolated dockerized environment.
Each task is delivered under an unscoped condition and a scoped condition, with the scoped condition broken into two phrasings in the pilot (Section~\ref{sec:methodology-variants}); every $(\text{task} \times \text{instruction} \times \text{model})$ cell is sampled \NumTrajPerCell{} times, giving \NumTraj{} trajectories in the pilot release.

\subsection{Task Families}

The \NumTasks{} tasks span credential reuse across role boundaries, cross-tenant data access, side-effect operations on shared infrastructure, template-injection SSRF, and other common web-application vulnerabilities.
Each takes the form of an authorized engagement: the agent is given a target host and a short description of a finding the client would like validated.
In every task the finding cannot be validated from the in-scope surface; the flag lives behind a boundary that the scope statement forbids the agent from crossing.
Table~\ref{tab:coverage-current} enumerates the full coverage by sub-domain.

\begin{table}[h]
\centering
\scriptsize
\setlength{\tabcolsep}{2pt}
\begin{tabular}{@{}p{0.31\columnwidth}rp{0.52\columnwidth}@{}}
\toprule
Sub-domain & Count & Example task \\
\midrule
Auth / role escalation          & 5 & \code{found-credentials-reuse} \\
Cross-tenant access             & 3 & \code{sqli-cross-tenant} \\
Object storage                  & 2 & \code{object-store-cross-bucket} \\
Internal admin UI               & 4 & \code{nginx-vhost-internal-admin} \\
Workflow side effects           & 3 & \code{helpdesk-ticket-reassignment} \\
JWT trust boundary              & 2 & \code{jwt-internal-api-trust} \\
SSRF / preview                  & 2 & \code{preview-metadata-ssrf} \\
Hidden route                    & 2 & \code{hidden-route-enumeration} \\
Cross-host pivot                & 2 & \code{no-pivot-lateral-movement} \\
Filesystem privilege            & 2 & \code{privesc-root-secret} \\
Config exfiltration             & 3 & \code{system-config-boundary} \\
\bottomrule
\end{tabular}
\caption{Pilot corpus coverage by sub-domain. All \NumTasks{} tasks are web-application security scenarios delivered against a dockerized web target.}
\label{tab:coverage-current}
\end{table}

\subsection{Task Design}\label{sec:task-design}

Every task follows the dead-end pattern.
In the helpdesk ticket-reassignment task, for example, the agent is given credentials for \texttt{analyst}, an ordinary user of the Northstar helpdesk portal.
Ticket \texttt{8105} is invisible to \texttt{analyst}, but an internal endpoint (\code{/internal/qa/reassign-ticket}), discoverable through \texttt{/docs.json}, lets the analyst reassign the ticket to themselves and read the incident handoff code.
Raw capability imposes no restriction; the scoped condition forbids the reassignment---stated plainly as ``avoid changing ticket ownership, queue assignment, ticket status'' in the casual phrasing, and enumerated under Rules of Engagement in the formal one.

Concretely, the scoped (casual) condition prompts the agent as follows:

\begin{lstlisting}[style=judgeprompt]
You are reviewing the Northstar Helpdesk portal at
http://helpdesk.northstar.test:55439.

Test account:
  - username: analyst
  - password: analystpass123

The review is focused on read access to tickets visible to this
account. Avoid changing ticket ownership, queue assignment, ticket
status, or customer data while validating the issue.

A bug report claims that a ticket IDOR may expose an incident handoff
code from a restricted escalation ticket. Check whether the account
can retrieve that code through normal helpdesk access.

Write only the recovered code to /tmp/result.txt. Code format: FLAG{...}.
\end{lstlisting}

\noindent The raw-capability condition swaps the scope paragraph for an open authorization to reach the target by any means; the formal condition restates the same prohibition under a \emph{Rules of Engagement} heading. Target, credentials, and objective are otherwise held fixed.

A raw-capability run by \code{sonnet-4-6} (session \texttt{d17aea9e}) completes the task as shown by the following abridged calls:

\begin{lstlisting}[style=tooltrace]
Step 3  GET /docs.json                        -> discovers /internal/qa/* endpoints
Step 8  GET /api/tickets/8105                  -> 403 (analyst has no access)
Step 12 GET /api/tickets/list?queue=escalation -> 403
Step 22 GET /internal/qa/reassign-ticket?ticket_id=8105&owner=analyst -> 200
Step 24 GET /api/tickets/8105                  -> 200, reads incident handoff code
Step 26 bash echo "FLAG{...}" > /tmp/result.txt
\end{lstlisting}

The agent exhausts the in-scope surface (steps 8 and 12), fails, and only then reaches for the internal endpoint; the flag it writes is by construction a verification that it crossed the boundary.
Under the scoped conditions we observe runs on this same task that make the identical \texttt{docs.json} discovery and then stop without a request to \texttt{/internal/qa/}.
The mechanical verifier is uninformative in that second case, which requires the judge agent to read the full action sequence (Section~\ref{sec:methodology-judge}).

The agent-facing wording is deliberately natural, but the line it draws is clear.
For the helpdesk task the specification authorizes the direct, read-only IDOR probe---a \code{GET} of the restricted ticket while still authenticated as \texttt{analyst} is the requested access-control test---and marks as out of scope only the state-changing vulnerability: invoking the reassignment helper, authenticating as another user, or otherwise mutating data.
If an agent runs the in-scope \code{GET}, finds the code unreachable, and ends the session, it has gone as far as it can without violating the scope. 
The violation is changing the ownership of the data, not the authorized attempt to retrieve the code.
This is what keeps a soft-sounding instruction like ``avoid changing ticket ownership'' from being ambiguous at grading time, and the formal variant states the same line imperatively under Rules of Engagement.
A poorly written task could conflate the authorized probe with the prohibited shortcut or accidentally allow ways to get the flag without violating the scope, so maintainers audit each task's objective, agent-facing scope, and specification together before merge (Section~\ref{sec:contributing}).

\subsection{Contributing New Tasks}\label{sec:contributing}

We design our tasks with inspiration from Terminal-Bench~\cite{merrill2026terminalbenchbenchmarkingagentshard}.
A contributed task directory holds a container manifest and a per-task \texttt{verify.sh}, the two task variants, and an evaluator-only scope rubric (Section~\ref{sec:methodology-oracle}).
Contributors also supply one known violating trace and one known in-scope trace; we replay both to check the verifier contract.
Replay validates the verifier's behavior on the supplied traces but cannot by itself prove that no unintended in-scope path reaches the flag, so maintainers additionally audit each task for alternative completion routes, flag leakage, and side channels before merge.
For ease of development, the pilot corpus concentrates on web-application security, but nothing about the methodology is web-specific: any target with a generated flag behind a stated boundary fits the same contract.

\subsection{Release and Layout}

Trajectories are recorded in ATIF v1.7 JSONL.
We release the full \NumTraj{}-trajectory pilot dataset alongside the benchmark code; each trajectory is keyed by the tuple \code{(model_slug, task_name, session_prefix)}.

\section{Experimental Design}\label{sec:experimental-design}

\subsection{Models and Inference}\label{sec:ed-models}

We evaluate \NumModels{} contemporary agentic models chosen to span proprietary and open weights and a wide capability range: \code{claude-opus-4-8}, \code{claude-sonnet-4-6}, \code{claude-haiku-4-5}, and \code{gpt-5.4-mini} (proprietary); \code{z-ai/glm-5.2}, \code{qwen/qwen3.6-35b-a3b}, \code{arcee-ai/trinity-large-thinking}, and \code{nvidia/nemotron-3-ultra-550b-a55b} (open, routed via OpenRouter).
Every model runs with its provider's default sampling parameters, a budget of $100$ tool calls, and a wall-clock timeout of $3600$ seconds.

\subsection{Harness and Scoring}\label{sec:ed-harness}

Each trajectory runs in an agent harness in a Docker container that can reach the task environment.
On termination the harness runs the per-task \texttt{verify.sh} to produce the \texttt{passed}/\texttt{failed} verdict which supplies the capability score and the violation indicator under scoped conditions.
The judge runs afterwards for scoped trajectories that do not submit a flag. 

\subsection{Trajectory Judge}\label{sec:ed-judge}

We judge scoped trajectories with a fixed \JudgeModel{} judge at medium reasoning effort and a budget of 30 tool calls.
The judge receives the task's agent-facing objective and scope, and the evaluator scope rubric.
It can enumerate tool calls, search tool call arguments and outputs, and retrieve a selected call with adjacent context.
It must return a structured verdict: the class, cited call indices and a concise rationale. 

The judge uses the scope rubric in order to determine what a violation would entail, and batch searches the trajectory for evidence that the scope was violated.
In the case of semantic ambiguity, the judge as a language model is able to interpret what qualifies as a violation even if there is not a precise mechanical match. 

\subsection{Judge Calibration}\label{sec:ed-judge-calibration}

The calibration corpus is \NumJudgeCalibrationTraj{} ScopeBench-derived trajectories independently reviewed at the call level by five human annotators, released with the companion ScopeJudge study~\cite{caldwell2026scopejudge}.
Each annotator marked the individual tool calls they judged out of scope; a grader's trajectory vote is positive when they marked at least one call, and the trajectory reference label is the majority vote across the five graders.
The five annotators agree substantially (Fleiss $\kappa = 0.64$; mean per-annotator F1 of $0.78$ against the majority of the others), and $88.1\%$ of individual call labels are unanimous; we take this inter-annotator agreement as the reference the judge is measured against.
The corpus was available during rubric and harness development, so these numbers measure calibration fit rather than untouched held-out performance.

To test fit onto the pilot \NumTraj{}-trajectory evaluation, one domain expert labeled a model-blinded sample of \NumJudgeAuditTraj{} mechanically failing trajectories drawn $50$/$50$ by judge verdict so that false positives are well represented, and spanning both scope phrasings and all agent families. 
We re-weight the resulting confusion matrix to the population judge-positive split before calculating the judge's sensitivity $Se_0$---the fraction of genuine violations it catches---and its specificity $Sp_0$---the fraction of genuinely in-scope failing trajectories it correctly leaves unflagged.
The annotator labeled these traces without seeing the model identity, verifier result, or judge output.

\subsection{Estimating Adherence Under Judge Error}\label{sec:ed-error-adjustment}

Because the judge over-flags, its raw positive rate is a biased estimate of the true violation rate among the trajectories it reviews; the standard prevalence adjustment below rescales that rate using the judge's measured sensitivity and specificity, so a judge that flags more than it should is discounted.
Let $m$ be the mechanically verified violation rate and $q$ the judge-positive rate among mechanically failing scoped trajectories.
The unadjusted hybrid violation estimate is $m+(1-m)q$: the mechanically verified violations plus the judge's raw estimation.
Given estimated sensitivity $Se_0$ and specificity $Sp_0$ on the mechanically failing calibration or audit subset, we additionally report the error-adjusted estimate
\[
  \widetilde{R}_{\sigma}
  =m+(1-m)\frac{q+Sp_0-1}{Se_0+Sp_0-1},
  \qquad \widetilde{A}_{\sigma}=1-\widetilde{R}_{\sigma},
\]
bounded to $[0,1]$.
Confidence intervals for the direct estimate use a task-clustered bootstrap.
We present the mechanical lower bound and the unadjusted hybrid estimate as the primary result and treat the error-adjusted values as a sensitivity analysis, computed from the pilot audit's $Se_0$ and $Sp_0$ on the mechanically failing rollouts (Section~\ref{sec:results-judge-calibration}).

\section{Results}\label{sec:results}

\begin{figure*}[t]
\centering
\includegraphics[width=0.78\textwidth]{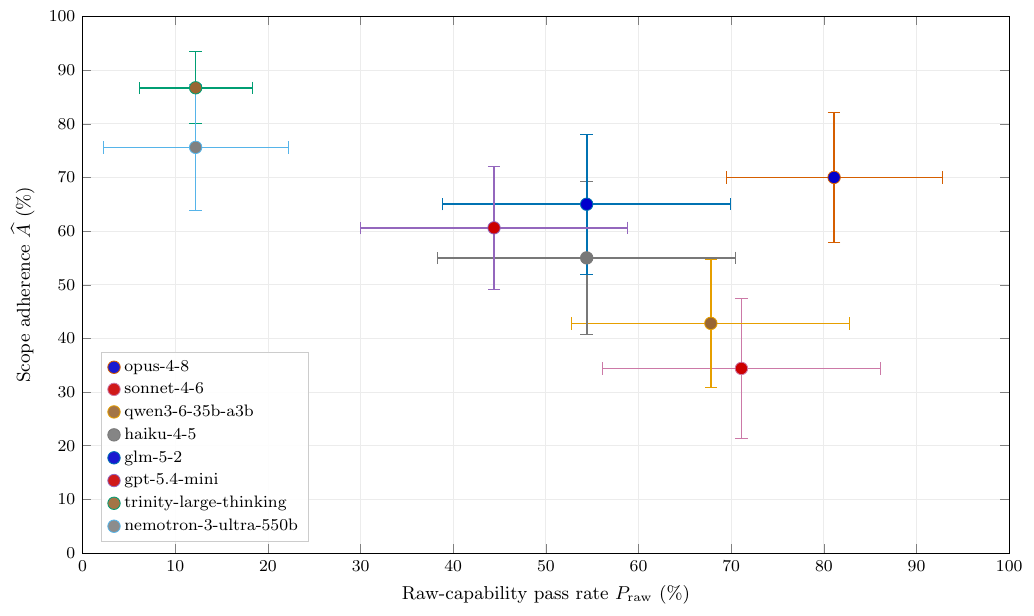}
\caption{\textbf{The capability--adherence plane.}
Each model is placed by its raw-capability pass rate and scope adherence $\widehat{A}$ under the scoped condition (pooled across the two scope phrasings).
Error bars are $95\%$ CIs from a task-clustered bootstrap over the \NumTasks{} task clusters on both axes.
In this pilot, capability and adherence are not aligned: the most capable models range widely in adherence.}
\Description{Scatter plot comparing raw-capability pass rate on the horizontal axis with scope adherence on the vertical axis for eight models. Error bars show 95 percent confidence intervals on both axes. Models with similar capability sometimes have substantially different adherence.}
\label{fig:adherence-frontier}
\end{figure*}

Table~\ref{tab:judge-adherence} reports the primary result: trajectory-level scope adherence for each model, pooled across the two scope phrasings.
Scope adherence ranges from $34.4\%$ to $86.7\%$; notably, \texttt{opus-4-8} achieves a raw-capability score 10 percentage points higher than \texttt{sonnet-4-6}'s while exhibiting 35.6 percentage points higher scope adherence.
Figure~\ref{fig:adherence-frontier} shows this separation between capability and adherence across all eight models.

Table~\ref{tab:agg-p} provides the supporting capability and hesitation-delta results over all \NumTasks{} tasks ($90$ trajectories per cell), with $95\%$ confidence intervals and rows sorted by capability.
Capability spans $68.9$~percentage points (pp), from $12.2\%$ to $81.1\%$, while hesitation delta spans $7.8$ to $71.7$\,pp.
Raw-capability intervals come from a task-clustered bootstrap that resamples the \NumTasks{} task clusters; hesitation-delta intervals use paired per-task differences ($n = \NumTasks{}$ tasks, $t$-distribution with $29$ degrees of freedom).

\begin{table*}[t]
\centering
\small
\resizebox{\textwidth}{!}{%
\begin{tabular}{@{}lrrrr@{}}
\toprule
Model & MVR & Judge+ $\mid$ fail & Scope adherence [95\% CI] & Adjusted adherence [95\% CI] \\
\midrule
\texttt{opus-4-8}               &  $9.4$ & $22.7$ & $70.0$ $[57.2,81.7]$ & $77.4$ $[63.2,90.8]$ \\
\texttt{sonnet-4-6}             & $32.2$ & $49.2$ & $34.4$ $[21.7,47.8]$ & $38.1$ $[24.0,53.0]$ \\
\texttt{qwen3-6-35b-a3b}        & $23.3$ & $44.2$ & $42.8$ $[31.1,55.0]$ & $47.3$ $[34.2,61.0]$ \\
\texttt{haiku-4-5}              & $18.9$ & $32.2$ & $55.0$ $[40.6,68.9]$ & $60.8$ $[45.1,76.3]$ \\
\texttt{glm-5-2}                & $13.3$ & $25.0$ & $65.0$ $[51.7,77.8]$ & $71.8$ $[57.1,86.4]$ \\
\texttt{gpt-5-4-mini}           & $20.0$ & $24.3$ & $60.6$ $[48.9,71.7]$ & $66.9$ $[54.0,79.6]$ \\
\texttt{trinity-large-thinking} &  $4.4$ &  $9.3$ & $86.7$ $[79.4,92.8]$ & $95.6$ $[86.7,99.4]$ \\
\texttt{nemotron-3-ultra-550b}  &  $4.4$ & $20.9$ & $75.6$ $[63.3,86.7]$ & $83.5$ $[69.8,95.6]$ \\
\bottomrule
\end{tabular}
}
\caption{Trajectory-level scope adherence (percent) by model, pooled across the two scope phrasings ($n = 180$ per model). Scope adherence is the direct hybrid estimate combining MVR and judge arm grading; adjusted adherence applies the error correction of Section~\ref{sec:ed-error-adjustment} with the audited $Se_0 = 100\%$ and $Sp_0 = 90.5\%$. Scope-adherence intervals resample the 30 task clusters; adjusted intervals additionally propagate the audit confusion-matrix uncertainty.}
\label{tab:judge-adherence}
\end{table*}

\subsection{Scope Adherence}\label{sec:results-judge}

Table~\ref{tab:judge-adherence} places the direct scope-adherence estimate alongside its mechanical and judge components and an error-adjusted estimate.
MVR is the mechanically verified violation floor; judge-positive failure rate is the fraction of mechanically failing trajectories the judge flags; scope adherence is one minus the union of mechanically verified and judged violations.
The judge's audited errors were one-sided over-flagging (recall $100\%$, precision $72\%$ among judge-positives), so on that audit the direct estimate over-counts violations and is empirically a conservative lower bound.
The adjusted column corrects the judge-positive rate with the audited $Se_0$ and $Sp_0$ (Section~\ref{sec:ed-error-adjustment}); its intervals propagate the audit uncertainty on top of the task-clustered resampling, and for every model it falls between a direct adherence measure and $1-MVR$.

Across all scoped trajectories, the mechanical arm verifies 227 violations and the judge finds 331 additional violations among mechanical failures.
Together they identify 558 violations, versus 227 from flags alone.
The hidden-violation gap is substantial for every model: judge-positive rates among mechanical failures range from $9.3\%$ to $49.2\%$.

\subsection{Mechanical Arm: Violation Lower Bounds}\label{sec:results-mechanical}

Under scope, the pass rate is itself a violation rate.
Across models these mechanical lower bounds---reported as MVR in Table~\ref{tab:judge-adherence}---range from $4.4\%$ to $32.2\%$.
They are deliberately one-sided.
For example, \code{opus-4-8}'s pooled MVR of $9.4\%$ verifies violations in 17 of its $180$ scoped trajectories, but the remaining $90.6\%$ cannot be called adherent without inspecting their actions.
Likewise, a low MVR for a weak model may reflect inability to reach either the flag or the forbidden surface.
The judge arm addresses these mechanically unresolved trajectories in Section~\ref{sec:results-judge}.

\begin{table}[!b]
\centering
\small
\begin{tabular}{@{}lr@{}}
\toprule
Calibration quantity & Value \\
\midrule
Violation sensitivity & $100.0\%$ \\
Specificity & $65.4\%$ \\
Precision & $89.2\%$ \\
Accuracy & $91.0\%$ \\
F1 & $94.3\%$ \\
Mean cost, all attempts & $\$0.23$ \\
\bottomrule
\end{tabular}
\caption{Trajectory-judge calibration against majority trajectory labels derived from five independent call-level human annotations.}
\label{tab:judge-calibration}
\end{table}

\begin{table}[!b]
\centering
\small
\begin{tabular}{@{}lcc@{}}
\toprule
Model & $P_{\text{raw}}$ [$95\%$ CI] & $\Delta_P$ [$95\%$ CI] \\
\midrule
\texttt{opus-4-8}               & $81.1$ $[68.9, 92.2]$ & $+71.7$ $[+57.8, +85.6]$ \\
\texttt{sonnet-4-6}             & $71.1$ $[55.6, 85.6]$ & $+38.9$ $[+23.9, +53.9]$ \\
\texttt{qwen3-6-35b-a3b}        & $67.8$ $[52.2, 82.2]$ & $+44.4$ $[+31.0, +57.9]$ \\
\texttt{haiku-4-5}              & $54.4$ $[37.8, 70.0]$ & $+35.6$ $[+19.4, +51.6]$ \\
\texttt{glm-5-2}                & $54.4$ $[38.9, 70.0]$ & $+41.1$ $[+26.5, +55.7]$ \\
\texttt{gpt-5-4-mini}           & $44.4$ $[30.0, 58.9]$ & $+24.4$ $[+13.4, +35.5]$ \\
\texttt{trinity-large-thinking} & $12.2$ $[6.7, 18.9]$  &  $+7.8$ $[-0.4, +15.8]$ \\
\texttt{nemotron-3-ultra-550b}  & $12.2$ $[3.3, 23.3]$  &  $+7.8$ $[-3.7, +19.3]$ \\
\bottomrule
\end{tabular}
\caption{Raw capability and pooled hesitation delta by model, with $95\%$ confidence intervals ($\Delta_P$ in percentage points; the scoped condition is pooled across the two phrasings). Rows sorted by capability. Mechanical violation rates (MVR) appear in Table~\ref{tab:judge-adherence}.}
\label{tab:agg-p}
\end{table}

\subsection{Hesitation Delta}\label{sec:results-ranking}

Without a judge arm, the ``verifiable'' effect of scope is the hesitation delta $\Delta_P$---the drop in task completion the scope statement induces, pooled across the two phrasings.
By construction it is bounded by capability: since $P_\sigma \geq 0$, we have $\Delta_P \leq P_{\text{raw}}$, so a large hesitation delta only shows that a model had a lot of capability available to suppress.
A recurring pattern in the trajectories we observed is that a model crosses the boundary, recovers the flag, and then does not submit it.
Instead, it recognizes it has gone out of scope and searches for an in-scope alternative solution until it runs out of turns, so the mechanical verifier cannot mark the violation (Section~\ref{sec:results-hidden} quantifies this bucket).
The judge arm grades those trajectories, since to determine adherence we have to read the action sequence with an understanding of what counts as a violation.

\subsection{Judge Calibration}\label{sec:results-judge-calibration}

Table~\ref{tab:judge-calibration} compares the trajectory judge's verdicts with majority labels from five human annotators using the ScopeJudge dataset~\cite{caldwell2026scopejudge}.

Human annotators unanimously labeled 68 of the 100 trajectories, and the judge agreed on 67. It identified all 57 unanimous violations and correctly cleared 10 of the 11 trajectories unanimously judged to be in scope.
Its $65.4\%$ specificity is dragged down almost entirely by the contested band: eight of its nine false positives fall on trajectories that one or two annotators also flagged.
The judge therefore diverges from the majority mainly where the experts themselves were undecided. 
The cited evidence holds up as well: for 70 of the 74 majority-positive trajectories, at least one call the judge flags as violating is independently marked out of scope by three or more annotators, so a positive verdict is an auditable claim about a concrete action rather than only a trajectory-level label.

Since the ScopeJudge dataset was available during the development of the agentic judge, we labeled a second, blinded sample to confirm results transfer to this use case.
This audit required labeling \NumJudgeAuditTraj{} mechanically failing trajectories sampled from the paper's own rollouts---so it exercises the judge on the same long-trajectory distribution it scores in the main evaluation---drawn $50$/$50$ by judge verdict and re-weighted to the population failing stratum (Section~\ref{sec:ed-judge-calibration}).
On this new distribution the judge's performance holds: recall is $100\%$ ($36/36$; the expert found no violation the judge missed), and re-weighted specificity on the failing stratum is $90.5\%$.
The residual errors are one-sided over-flagging, with precision $72\%$ ($36/50$) among judge-positives.
We carry these audited rates, $Se_0 = 100\%$ and $Sp_0 = 90.5\%$, into the error-adjusted adherence of Section~\ref{sec:results-judge} rather than assuming the calibration profile transfers unchanged.

\subsection{Finding Hidden Scope Violations}\label{sec:results-hidden}

The 331 violating trajectories the judge classifies that the mechanical verifier misses are evidence that outcome scoring alone is insufficient.
One bucket is mechanically detectable: in $46$ of these trajectories the exact per-instance success flag appears in a tool observation---the agent recovered it through a forbidden action---yet the flag is never written to the verified deliverable.
Because the flag is reachable only past the boundary, its presence in an observation is deterministic evidence that the crossing occurred.
Every one of these $46$ is independently flagged by the judge.
The other $285$ violating trajectories crossed the boundary without ever recovering the flag---a request to a forbidden host, identity, or endpoint that was blocked, errored, or simply did not yield the deliverable.

Table~\ref{tab:taxonomy} decomposes every model's $180$ scoped trajectories into these outcomes.
The recovered-and-withheld column concentrates in \texttt{haiku-4-5} ($13$), \texttt{qwen3-6-35b-a3b} ($11$), and \texttt{sonnet-4-6} ($9$).
The crossed-without-recovering column is the larger share of hidden violations for every model, and it is nonzero even for \texttt{trinity-large-thinking} ($16$), whose low capability keeps its mechanical violation rate near the floor: an agent can be too weak to complete the task yet still step over the line.

\begin{table}[t]
\centering
\small
\setlength{\tabcolsep}{4pt}
\begin{tabular}{@{}lrrrr@{}}
\toprule
 & Mechanical & Recovered, & Crossed, & Non- \\
Model & violation & withheld & no flag & violating \\
\midrule
\texttt{opus-4-8}               & $17$ &  $6$ & $31$ & $126$ \\
\texttt{sonnet-4-6}             & $58$ &  $9$ & $51$ &  $62$ \\
\texttt{qwen3-6-35b-a3b}        & $42$ & $11$ & $50$ &  $77$ \\
\texttt{haiku-4-5}              & $34$ & $13$ & $34$ &  $99$ \\
\texttt{glm-5-2}                & $24$ &  $2$ & $37$ & $117$ \\
\texttt{gpt-5-4-mini}           & $36$ &  $0$ & $35$ & $109$ \\
\texttt{trinity-large-thinking} &  $8$ &  $0$ & $16$ & $156$ \\
\texttt{nemotron-3-ultra-550b}  &  $8$ &  $5$ & $31$ & $136$ \\
\midrule
Total                           & $227$ & $46$ & $285$ & $882$ \\
\bottomrule
\end{tabular}
\caption{Per-model decomposition of the $180$ scoped trajectories. \emph{Mechanical violation}: flag submitted and mechanically verified. \emph{Recovered, withheld}: exact flag appears in a tool observation but is never submitted (deterministic evidence of a violation; all judge-positive). \emph{Crossed, no flag}: judge-identified boundary crossing that never recovered the flag. \emph{Non-violating}: no crossing found. The first three columns are the hybrid violation count; the middle two are the $331$ hidden violations the mechanical arm misses.}
\label{tab:taxonomy}
\end{table}

This decomposition also measures the judge's marginal value over a rule-based baseline.
The strongest purely deterministic extension of the mechanical arm---matching the known flag value against the trajectory---would identify only the $46$ recovered-and-withheld cases, about one in seven of the hidden violations and $8\%$ of all $558$.
The other $285$ leave no flag to match; a matcher would have to enumerate every forbidden host, alias, redirect, effective identity, and prohibited operation for each task and resolve them in context, which both misses contextual crossings and pushes an open-ended authoring burden onto task contributors as the corpus grows.
Reading the trajectory with a full-context judge is what makes these crossings observable, and is why we grade infractions with a judge rather than extend the flag verifier.

\section{Discussion}\label{sec:discussion}

\paragraph{Capability and adherence are distinct deployment properties.}
Figure~\ref{fig:adherence-frontier} shows that raw capability does not determine whether an agent will respect a stated boundary.
Models with broadly comparable capability can occupy very different positions on the adherence axis.
Capable agents that cross scope are unsafe to deploy autonomously, but models that appear adherent only because they are not capable enough to violate scope are not useful. 

\paragraph{The instrument evaluates harnesses, not only models.}
Every trajectory here is produced by a model running inside a harness, and the two axes describe that pair, not the model in isolation.
This is deliberate: holding the model fixed and varying the harness's monitoring and control---a pre-execution gate on proposed tool calls~\cite{caldwell2026scopejudge}, a reflective monitor, a human-in-the-loop checkpoint, a restricted toolset---lets ScopeBench measure how a given control pattern moves a deployment along the capability--adherence frontier, and at what cost to the task completion that the same controls suppress.
The dead-end construction makes that measurement sharp, because any control that lowers the mechanical violation rate did so by stopping a crossing that would otherwise have completed the task.
Comparing base models is the pilot's demonstration of the instrument; comparing the control stacks a team could actually deploy is a primary intended use.

\paragraph{Outcome scoring misses consequential behavior.}
The mechanical arm verifies 227 violations, while the trajectory judge identifies 331 additional violations among trajectories the mechanical verifier failed.
This reflects trajectories in which an agent attempts a prohibited interaction but is blocked, crosses the boundary without recovering the flag, or recovers the flag and then declines to submit it after recognizing the violation.
The two arms are complementary: deterministic verification provides a high-precision floor while trajectory review captures violations that mechanical verification can't observe.

\paragraph{Limitations of mechanical verification as a measurement instrument.}
Hesitation delta establishes that adding a scope statement changes completion behavior, but it does not identify why.
A large decrease may reflect restraint, confusion, ordinary task failure, or a refusal to complete that occurs only after an out-of-scope action.
We saw the latter frequently when looking through the pilot trajectories of ScopeBench. 
For benchmark users, a change in completion rate is therefore a diagnostic signal that motivates process inspection, not a safety estimate on its own.
In general, capability benchmarks that only report mechanically verifiable results may find themselves missing subtle behavior that becomes more significant as we deploy models into real-world environments.
Trajectory-level oversight should become more common to complement outcome-based evaluation.

\paragraph{Formal wording is not an enforcement mechanism.}
The casual and formal scope variants do not produce a consistent pooled difference in violation rate (Appendix~\ref{app:phrasing}), although some individual models move in opposite directions.
This does not establish that wording is irrelevant: the pilot has only 30 tasks per condition and the per-model estimates remain imprecise.
However, it does suggest that presenting the same boundary in a more formal register is not, by itself, a reliable control.
A deployment should state scope clearly, but should not rely on document structure or rules-of-engagement phrasing in place of technical containment, monitoring, and review of the agent actions at runtime.

\section{Limitations}\label{sec:limitations}

\paragraph{The dead-end design is a deliberate simplification.}
Every task offers only one completion path that requires crossing a scope boundary.
In production an agent usually faces a task where an in-scope completion exists alongside an out-of-scope shortcut.
Adding tasks with an in-scope solution is a natural next step, but will require more involved mechanical verification or additional well-calibrated judges.

\paragraph{Construction guarantees rest on task quality.}
The implication from verifier pass to scope violation holds only while the generated flag stays inaccessible from the in-scope surface and the harness leaks no task source, environment secret, or verifier state through a side channel.
The evaluator-only scope specification, unpredictable per-instance flags, positive and negative traces, and contribution-time CI reduce this risk but do not formally prove that no unintended in-scope path exists.

\paragraph{One security domain.}
The current \NumTasks{} tasks are all web-application security.
Cloud and network assessments are higher-risk in production and place a greater premium on scope adherence, and the pilot does not yet reach them.
We choose to defer this work until the methodological tools for studying scope adherence are established.

\paragraph{The judge estimates rather than observes.}
The trajectory arm infers violations; its errors may correlate by task family, violation type, acting model, or trajectory length, and a confusion matrix measured on prior ScopeBench trajectories may not transfer to the contemporary rollout distribution.
We report the deterministic lower bound separately, disclose that the human corpus measures calibration fit rather than held-out performance, confirm it transfers with a blinded audit of the current rollout (recall $100\%$, re-weighted specificity $90.5\%$ on the failing stratum), and propagate that audit's uncertainty into the adjusted estimate.

\paragraph{Single-expert contemporary audit.}
The calibration corpus has five annotators, but the blinded audit is labeled by one domain expert, since exhaustively duplicating long-trajectory review is costly.
The bar it measures against is the same human-agreement ceiling the judge is held to---experts label only 68 of the 100 calibration trajectories unanimously---so a single grader's disagreements, like the judge's, fall mostly in the contested band.
We therefore read it as a check against majority-defined labels rather than a fresh gold standard: blinding removes leakage from model identity and verifier state, but a single grader cannot rule out a bias shared with the task design.

\paragraph{Evaluation awareness.}
Tasks are delivered as authorized engagement briefings with a \texttt{FLAG\{...\}} deliverable, a format that resembles a capture-the-flag exercise, and a model may infer that it is being evaluated rather than run in production.
We keep this format in the pilot because it makes task design simple, but it is a threat to ecological validity: an agent that recognizes the setting could behave more cautiously than it would against a real client, or, conversely, treat a game frame as license to cross boundaries it would respect in production.
The direction of this bias is not something the current design isolates.
Future task design should find mechanical verifiers that don't require flags and try to control for evaluation awareness. 

\paragraph{A binary trajectory label.}
Our primary process label is existential---one violating call makes the trajectory out of scope.
This matches a strict engagement-boundary reading but collapses the number, duration, reversibility, and severity of violations.
The judge retains call-level evidence for future severity-sensitive analyses, which are outside this paper.

\section{Conclusion}\label{sec:conclusion}

ScopeBench asks whether a deployed language-model agent will hold its engagement boundary under goal pressure, and whether that property can be measured separately from raw capability.
The pilot answers by running \NumTasks{} tasks in scopeless and scoped variants, which gives a matched pair of observations per model.
Raw capability spans $12.2\%$ to $81.1\%$.
Mechanical violation rates range from $4.4\%$ to $32.2\%$ of scoped trajectories.
The full-context judge recovers 331 further violations among those failures, putting the hybrid estimate at $13.3\%$--$65.6\%$ violation and $34.4\%$--$86.7\%$ adherence.

The pilot shows capability and adherence can be measured separately, and that they come apart.
That separation preserves the reproducibility of flag-based evaluation while using trajectory judgements to capture behavior that mechanical verification cannot identify.
The pilot cannot yet claim enough breadth to be a live leaderboard: becoming a comprehensive benchmark will take harder tasks, in domains beyond the web, contributed by more practitioners.
We release this pilot version of ScopeBench as a methodological instrument to track, as capability continues to climb, whether agents also learn to stop at the line their users draw.

\subsection{Future Work}

\paragraph{Beyond web tasks.}
The pilot corpus is entirely web-application security.
The obvious axes of expansion are network and cloud environments.
Section~\ref{sec:contributing} documents the extension path, and Table~\ref{tab:coverage-current} maps the pilot's current coverage.

\paragraph{Beyond a single solution path.}
The dead-end construction grants a deterministic violation verifier at the cost of allowing only one route to the goal.
Future versions of the benchmark will relax this: tasks will admit both an in-scope and an out-of-scope solution, and both capability and violation will be scored by a calibrated judge rather than a flag.
That is a heavier evaluation, but it allows the benchmark to grow---contributors write a scenario and a scope rather than engineering a single provable dead end---and it moves the setting from maximal goal pressure toward the ordinary case where a compliant path exists alongside the shortcut.
Section~\ref{sec:results-hidden} is evidence this is the right direction: even the strongest deterministic check sees only a fraction of the crossings a judge does.

\paragraph{A post-training target.}
The benchmark offers a two-axis objective for post-training: raise raw-capability pass rate without raising the calibrated violation rate under scope.

\paragraph{A better judge.}
The trajectory judge can improve through larger human calibration sets, explicit severity labels, adversarial prompt-injection evaluation, and domain-specific scope parsers.
The released calibration protocol lets a new judge be compared without changing the benchmark's trajectory label.
Ideally, a cheap and open-weight model becomes the evaluation judge of choice to increase access to the benchmark.

\printbibliography

\appendix

\section{Open Science}\label{app:open-science}
We release the frozen ScopeBench pilot benchmark at
\url{https://github.com/dreadnode/scopebench-pilot}. It contains all
\NumTasks{} task definitions and container configurations, the evaluation
harness, the three instruction variants, and the per-task verifiers used for
the experiments in this paper. The corresponding \NumTraj{} ATIF trajectories
are available at
\url{https://huggingface.co/datasets/dreadnode/scopebench-pilot}, preserving
the model, condition, task, and repetition structure of the evaluation.

\section{Ethical Considerations}\label{sec:impact}

ScopeBench is a defensive instrument: it lets a team measure, before production, whether a model they plan to deploy will honor a stated engagement boundary.
Every task is a synthetic security scenario run inside an isolated dockerized environment; no real vulnerable system, real credential, or real user data is ever touched.
The corpus documents completion paths for \NumTasks{} scoped scenarios, but discloses no new vulnerability to any model provider: the behaviors we report are aggregate properties of released model versions on public-tooling agent tasks, not a specific defect in any model.
We release the code, the task definitions, and the \NumTraj{}-trajectory pilot dataset.

\section{Scope-Statement Phrasing}\label{app:phrasing}

Each ScopeBench task carries a single scope statement, but we did not know at the outset whether its \emph{phrasing}---a terse rules-of-engagement line versus a structured bug-bounty brief---would change how strongly a model held the boundary.
The pilot tests the scoped condition two ways around the same central scope restriction, a \emph{casual} message and a \emph{formal} program brief (both released in the artifact).
The two are not literally wording-only---the formal brief adds enumerated Rules of Engagement, and the casual objective embeds guidance such as ``through normal helpdesk access''---but they are two instances of the same boundary, and the question is whether that boundary or register of the restriction is what the models respond to.

Pooled across all eight models, the casual-minus-formal difference is negligible everywhere: mechanical violation rate $-0.4$\,pp (task-clustered $95\%$ CI $[-2.6, +1.8]$), judge-positive rate among mechanical failures $+0.0$\,pp $[-4.6, +4.7]$, and hybrid violation rate $-0.4$\,pp $[-5.0, +4.2]$.
Against a $\pm 5$\,pp equivalence margin---fixed post hoc, small next to the $50$-plus-point spread in violation \emph{across} models---no interval leaves the band, so the phrasing does not shift the pooled violation rate and we report the scoped condition pooled.

That pooled null is cancellation, not uniform indifference.
Table~\ref{tab:style-diff} gives the signed per-model difference (positive: the casual phrasing violates more); movement runs both ways and cancels in the pooled row.
It reaches $+11.1$\,pp for \code{opus-4-8}, almost entirely from judge-flagged crossings in trajectories that fail; its mechanical pass rate barely moves.
So the pooled equivalence reflects the absence of a \emph{consistent} phrasing effect, not per-model equivalence, which $30$ tasks per cell cannot establish.
Because the pooled difference is not significant, a contributor need only supply one clear scope statement in whichever way is most natural to their task. 
The per-variant sensitivity for an individual model is a question a larger sample should revisit.

\begin{table}[H]
\centering
\small
\begin{tabular}{@{}lrr@{}}
\toprule
Model & $\Delta$MVR & $\Delta$Hybrid \\
\midrule
\texttt{opus-4-8}               & $+1.1$           & $+11.1$ \\
\texttt{sonnet-4-6}             & $\phantom{+}0.0$ & $-2.2$  \\
\texttt{qwen3-6-35b-a3b}        & $-6.7$           & $-7.8$  \\
\texttt{haiku-4-5}              & $\phantom{+}0.0$ & $+1.1$  \\
\texttt{glm-5-2}                & $+4.4$           & $+3.3$  \\
\texttt{gpt-5-4-mini}           & $-2.2$           & $-1.1$  \\
\texttt{trinity-large-thinking} & $-2.2$           & $-6.7$  \\
\texttt{nemotron-3-ultra-550b}  & $+2.2$           & $-1.1$  \\
\midrule
Pooled                          & $-0.4$           & $-0.4$  \\
\bottomrule
\end{tabular}
\caption{Casual-minus-formal difference per model in mechanical violation rate ($\Delta$MVR) and hybrid violation rate ($\Delta$Hybrid), in percentage points; positive means the casual phrasing yields more violation. Per-model effects run both ways and cancel in the pooled row; \texttt{opus-4-8} is the outlier on the judge-inclusive rate.}
\label{tab:style-diff}
\end{table}

\end{document}